%% file: emnlp2021.tex
\pdfoutput=1

\documentclass[11pt]{article}

\usepackage{emnlp2021}

\usepackage{times}
\usepackage{latexsym}

\usepackage[T1]{fontenc}

\usepackage[utf8]{inputenc}

\usepackage{microtype}

\usepackage{color}
\usepackage{amsmath}

\usepackage{graphicx}
\usepackage{natbib}
\usepackage{caption}
\usepackage{amsfonts}
\usepackage{amssymb}
\usepackage{multirow}
\usepackage{adjustbox}
\usepackage[english]{babel} 
\usepackage{arydshln}
\usepackage[ruled,linesnumbered]{algorithm2e}
\usepackage{bbding}
\usepackage{CJKutf8}
\usepackage{booktabs}

\SetKwInOut{Parameter}{Parameters}

\title{FewCLUE: A Chinese Few-shot Learning Evaluation Benchmark}

\author{
\begin{tabular}{c} 
Liang Xu, Xiaojing Lu, Chenyang Yuan,  \\ Xuanwei Zhang, Huilin Xu, Hu Yuan, Guoao Wei, Xiang Pan,  \\ Xin Tian, Libo Qin, Hai Hu 
 \\
  CLUE team \\
   {\tt \ CLUE@CLUEbenchmarks.com} 
   \end{tabular}
}

\begin{document}
\maketitle
\input{Abstract.tex}

\section{Introduction}
\label{Introduction}
\input{Introduction.tex}

\section{Related Work}
\label{RelatedWork}
\input{RelatedWork.tex}
\section{Dataset Construction}
\label{DatasetConstruction}
\input{DatasetConstruction.tex}

\section{Tasks}
\label{Tasks}
\input{Tasks.tex}

\section{Experiments}
\label{experiments}

\input{Experiments.tex}

\section{Related Work}
\label{related work}

\input{RelatedWork.tex}
\section{Conclusion}
\label{conclusion}
\input{Conclusion.tex}


\bibliography{anthology,custom}
\bibliographystyle{acl_natbib}



\end{document}

%% file: Abstract.tex
\begin{abstract}

Pretrained Language Models (PLMs) have achieved tremendous success in natural language understanding tasks. 
While different learning schemes---fine-tuning, zero-shot, and few-shot learning---have been widely explored and compared for languages such as English, there is comparatively little work in Chinese to fairly and comprehensively evaluate and compare these methods and thus hinders cumulative progress.
In this paper, we introduce the Chinese Few-shot Learning Evaluation Benchmark (FewCLUE), the first comprehensive few-shot evaluation benchmark in Chinese. It includes nine tasks, ranging from single-sentence and sentence-pair classification tasks to machine reading comprehension tasks. 
We systematically evaluate five state-of-the-art (SOTA) few-shot learning methods (including PET, ADAPET, LM-BFF, P-tuning and EFL), and compare their performance with fine-tuning and zero-shot learning schemes on the newly constructed FewCLUE benchmark.  
Experimental results reveal that: 
1) The effect of different few-shot learning methods is sensitive to the pre-trained model to which the methods are applied; 2) PET and P-tuning achieve the best overall performance with RoBERTa and ERNIE respectively. 
Our benchmark is used in the few-shot learning contest of NLPCC 2021\footnote{http://tcci.ccf.org.cn/conference/2021/}.
In addition, we provide a user-friendly toolkit, as well as an online leaderboard to help facilitate further progress on Chinese few-shot learning. We provide a baseline performance on different learning methods, a reference for future research.

\end{abstract}

%% file: Introduction.tex
With the burst of pre-trained models (BERT~\citep{bert-2019}, RoBERTa~\citep{liu2019roberta} , ELECTRA~\citep{clark2020electric}, XLNet~\citep{zhilinyang2020xlnet} and the evolution of unified evaluation benchmarks (GLUE~\citep{wang2018glue} and SuperGLUE~\citep{wang2019superglue}), the research of Natural Language Processing (NLP) has obtained signiﬁcant breakthroughs.
Recently, CLUE~\citep{clue} is introduced to facilitate research of the Chinese NLP community, whose speakers account for one-fourth of the world's population. 

Though achieving remarkable success, most of the existing benchmarks rely on a considerable amount of labeled data, which is used for evaluating the model's ability in the standard supervision setting. 
However, lots of high-quality human annotation can be costly and time-consuming.
To address the issue, few-shot or zero-shot learning scheme~\citep{gpt3,2021-gao-lm-bff,schick-schutze-2021-PET,schick2020small,liu2021gpt} has gained increasing attention, which makes it possible to obtain considerable performance with only a handful of annotated training data. 
While the increasing attention it has received, progress on few-shot settings in Chinese NLP is still slow. 
One of the primary constraints is the lack of a unified benchmark for few-shot Chinese NLP, thus new methods cannot be easily compared and iteratively improved.

To address this problem and facilitate few-shot studies in the Chinese language, we introduce a few-shot Chinese Language Understanding Evaluation (FewCLUE) benchmark that consists of a collection of nine different natural language understanding tasks, including single-sentence classification task, sentence-pair classification task and machine reading comprehension tasks. 
In addition, we systematically evaluate five state-of-the-art models (PET~\citep{schick-schutze-2021-PET}, ADAPET~\citep{tam2021-ADAPET}, EFL~\citep{wang2021-EFL}, Ptuning~\citep{liu2021gpt} and EFL~\citep{wang2021-EFL}) on our newly proposed benchmark. 
Experimental results reveal:
1) The effect of different few-shot learning methods is related to the pre-trained model; 2) PET/P-tuning achieves better performance than others learning methods.



In summary, our contribution is three-fold:
\begin{itemize}
    \item To the best of our knowledge, we construct the first systematic and comprehensive Few-shot Chinese Language Understanding Evaluation benchmark and provide strong baselines and human evaluation.
     We believe the benchmark will facilitate the research on few-shot learning for Chinese NLU.
    \item We systematically evaluate and analyze various state-of-the-art few-shot models on our proposed unified benchmark. The empirical findings can shed a light on future research in Chinese few-shot learning. 
    Our results show that there is no single few-shot learning method that outperforms others for all pre-trained language models that we experimented with. However, P-tuning and PET yield the best overall performance on the 9 tasks, when applied to RoBERTa and ERNIE respectively. 
    \item We introduce a user-friendly toolkit, as well as an online leaderboard with an auto-evaluation system, supporting all our evaluation tasks and models, with which researchers can reproduce experimental results and compare the performance of different submitted models easily.
\end{itemize}
Our benchmark and code are available at
\url{https://github.com/CLUEbenchmark/FewCLUE}.

%% file: RelatedWork.tex
\subsection{Few Shot Learning Benchmarks}
In computer vision, few-shot benchmarks have been proposed to measure the few-shot learning ability of pretrained computer vision models and the related few-shot learning methods~\cite{chen_2019_CLOSERLOOKFEWSHOT}.  
In NLP, there are also some methods considering the few shot learning process and achieved some good performance in standard benchmark~\cite{floridi2020gpt}. 
FewGLUE~\cite{schick2020small}, is a few-shot learning dataset, consisting of a random selection of 32 training examples from the SuperGLUE~\citep{wang2019superglue} training sets and up to 20,000 unlabeled examples for each SuperGLUE task. We observe that recent research generates their datasets from GLUE~\citep{wang2018glue} or SuperGLUE~\citep{wang2019superglue}, in a different way.

The difference from the previous benchmark is that it is the first comprehensive few-shot learning evaluation benchmark in Chinese. Secondly, to be the best of our knowledge, there is no such comprehensive few-shot learning evaluation benchmark in other languages. It not only includes a complete training set and verification set, but also multiple training-validation-test sets, evaluation can be more stable. Each task contains a large number of unlabeled samples. While investigating few-shot learning, unsupervised or semi-supervised learning can also be evaluated simultaneously. For better comparison, we also evaluate the performance of zero-shot learning and human. Furthermore, it contains an additional private test set, which can be used for fair evaluation on the leaderboard.

\subsection{Few Shot Learning Methods}

In this section, we briefly review current few-shot learning strategies for PLMs. We explain the implementation details in section~\ref{sec:baseline:models} and section~\ref{sec:implementation}.

PET converts a task into a template-based cloze task~\citep{schick-schutze-2021-PET,schick2020small}. Compared with the direct fine-tuning of downstream tasks through BERT-based PLMs, PET produces considerable performance improvement in the few-shot scenario; for some tasks the performance increase can be as large as 10 points. 
While the templates in PET are manually constructed, 
researchers have also explored various ways of automated template-construction. For example, 
LM-BFF~\citep{2021-gao-lm-bff} uses the generative model T5~\citep{2019t5} to generate templates. Ptuning~\citep{liu2021gpt} builds continuous end-to-end templates which are not in the form of natural language. 
Furthermore, there are other works transforming tasks into natural language inference (EFL, \citet{wang2021-EFL}). 
One of our goals in this paper is to systematically compare the performance of these methods. 

%% file: DatasetConstruction.tex
\subsection{Task Selection}
The goal of FewCLUE is to provide a reliable evaluation for few-shot learning, discover the challenges, and provide a basis for further research in Chinese by creating such a systematic and comprehensive benchmark. The benchmark aims to investigate the NLU capabilities using a variety of tasks, e.g., text classification, natural language inference, text similarity, reading comprehension, etc. These tasks can be generalized into single-sentence, sentence-pair, and general reading comprehension tasks. To ensure the quality and coverage of the language understanding tasks, we select tasks using the following criteria:


\paragraph{Representative} Tasks should reflect real-world settings, e.g., classification with many labels, samples from real text data.

\paragraph{Challenging and discriminative}  We expect the tasks to be challenging, showing the gap between human and machines. Moreover, the benchmark ought to better discriminate the performance of different models, resulting in more discrete results on the leaderboard.

\paragraph{Consistent} We expect that the results of the same model on different train/dev splits to be consistent.

\subsection{Building the Benchmark}\label{subsec:building}


In order for our FewCLUE benchmark to maximally satisfy the needs of the NLP community, we first sent out a public survey to collect opinions and suggestions on the benchmark. 

\paragraph{Public survey} 
The feedback we received include: 1) tasks should be diverse, including single-sentence, sentence-pair classification, and machine comprehension tasks; 2) Each dataset typically contains 16 or 32 samples per class; 3) Each task include enough unlabelled samples for research in zero-shot and semi-supervised learning; etc. The benchmark is built to satisfy these features.

Next, we picked 6 tasks from the CLUE~\citep{clue} benchmark, and then added 3 new tasks. See Table~\ref{tab:tasks:description} for detailed descriptions.  

\paragraph{Data sampling:} we adopt multiple sampling strategies to generate samples from original datasets to reflect real scenarios. Specifically, we collect 32 samples for tasks with $ n\_label \le 3$.
Five tasks fall into this category: EPRSTMT, OCNLI, BUSTM, CSL and CLUEWSC.
For tasks where $3 < n\_label \le 20$, we sample 16 samples per class; TNEWS falls into this category. 
For tasks where $n\_label > 20$ we sample 8 samples per class; CSLDCP and IFLYTEK fall into this category. For CHID, since each sample has 7 candidate idiom and the correct answer can appear at any position (1st to 7th), we collect 6 samples for each position where the correct answer locates (7 positions, $6 * 7 = 42$ samples in total).

\paragraph{Multiple training/validation splits:} since training on small datasets might lead to fluctuating results, we provide multiple training/validation splits to alleviate the issue and make our benchmark more robust. Participants should train and evaluate each split and submit their results on the unique private test set.

\paragraph{Public test set:} each task comes with a public test set, on which people can conduct experiments and publish results for research purposes.

\paragraph{Unlabeled data:} up to 20k unlabeled samples are also provided in each task for potential research in unsupervised and semi-supervised learning.

Overall, we create a comprehensive Chinese few-shot benchmark containing nine tasks, including single-sentence,sentence-pair, and machine reading comprehension tasks. We adopt different sampling methods to reflect the real-world setting while maintaining the few-shot characteristic. Also, we provide multiple training and validation sets for each task to ensure the benchmark’s robustness and release an additional public test set for further research. Finally, we release a large number of unlabeled samples for unsupervised and semi-supervised learning. The details of each task in FewCLUE are shown in Table \ref{tab:tasks:description}

%% file: Tasks.tex
\begin{table*}[t]
\footnotesize
	\begin{adjustbox}{width=1.1\textwidth}
\begin{tabular}{llllllllll}\toprule
\hline
Corpus & Train & Dev & Test Pub & Test Priv & N Labels & Unlabeled & Task & Metric & Source \\ \midrule
\hline
 \multicolumn{10}{l}{Single Sentence Tasks} \\\midrule
 \hline
EPRSTMT & 32 & 32 & 610 & 753 & 2 & 19565 & SntmntAnalysis & Acc & E-CommrceReview \\
CSLDCP & 536 & 536 & 1784 & 2999 & 67 & 18111 & LongTextClassify & Acc & AcademicCNKI \\
TNEWS & 240 & 240 & 2010 & 1500 & 15 & 20000 & ShortTextClassify & Acc & NewsTitle \\
IFLYTEK & 928 & 690 & 1749 & 2279 & 119 & 7558 & LongTextClassify & Acc & AppDesc \\ \midrule
 \hline
 \multicolumn{10}{l}{Sentence Pair Tasks}\\ \midrule
\hline
OCNLI & 32 & 32 & 2520 & 3000 & 3 & 20000 & NLI & Acc & 5Genres \\
BUSTM & 32 & 32 & 1772 & 2000 & 2 & 4251 & SemanticSmlarty & Acc & AIVirtualAssistant \\ \midrule
 \hline
  \multicolumn{10}{l}{Reading Comprehension} \\\midrule
 \hline
CHID & 42 & 42 & 2002 & 2000 & 7 & 7585 & MultipleChoice,idiom & Acc & Novel,EssayNews \\
CSL & 32 & 32 & 2828 & 3000 & 2 & 19841 & KeywordRecogntn & Acc & AcademicCNKI \\
CLUEWSC & 32 & 32 & 976 & 290 & 2 & 0 & CorefResolution & Acc & ChineseFictionBooks\\\bottomrule
 \hline
\end{tabular}
	\end{adjustbox}
\caption{Task descriptions and statistics.}
\label{tab:tasks:description}
\end{table*}

The following sections will briefly introduce the nine tasks in FewCLUE. 



\subsection{Single Sentence Tasks}

\paragraph{EPRSTMT} (E-commerce Product Review Dataset for Sentiment Analysis), also known as EPR-sentiment, is a binary sentiment analysis dataset based on product reviews on e-commerce platform. Each sample is labelled as Positive or Negative. It collect by ICIP Lab of Beijing Normal University. We filtered, test and re-organized it to make it suitable forFewCLUE.

\begin{CJK*}{UTF8}{gbsn}
\paragraph{CSLDCP} (Chinese Science Literature discipline Classification) is a text classification dataset including abstracts from a variety of Chinese scientific papers, with 13 scientific disciplines(field), and each discipline contains multiple categories. These papers belong to domains ranging from social science to natural science, including 67 categories, e.g., “口腔医学” (Oral medicine), “社会学” (Sociology) and “机械工程” (Mechanical engineering).

\end{CJK*}

 \paragraph{TNEWS}
TouTiao Text Classification for News Titles \citep{clue} consists of Chinese news published by TouTiao before May 2018, 
with a total of 73,360 titles.
Each title is labeled with one of $15$ news categories (finance, technology, sports, etc.)
and the task is to predict which category the title belongs to. To make the dataset more discriminative, we use cross-validation to filter out some of the easy examples (see Section D Dataset Filtering in the Appendix for details). We then randomly shuffle and split the whole dataset into a training set, development set and test set.

\paragraph{IFLYTEK}
IFLYTEK \citep{2019iflytek} contains
17,332 app descriptions. The task is to assign each description into one of $119$ categories, such as food, car rental, education, etc. A data filtering technique similar to the one used for the TNEWS dataset has been applied.
  
\subsection{Sentence Pair Tasks}
\paragraph{OCNLI} Original Chinese Natural Language Inference (OCNLI, \citet{ocnli}) is collected closely following procedures of MNLI \citep{mnli}. OCNLI is composed of inference pairs from five genres: news, government, fiction, TV transcripts and Telephone transcripts, where the premises are collected from Chinese sources, and universities students in language majors are hired to write the hypotheses. The annotator agreement is on par with MNLI. We believe the non-translation nature of OCNLI makes it more suitable than XNLI \citep{conneau2018xnli} as an NLU task specific for Chinese. 

\paragraph{BUSTM} (XiaoBu Dialogue Short Text Matching)~\citep{BUSTM} is a dataset provided by XiaoBu Assistant (Breeno), voice assistance produced by OPPO for its smartphones and IoT devices. The dataset comes from the data used to train XiaoBu on intent recognition through short text matching and mainly consists of colloquial sentence pairs.

\subsection{Machine Reading Comprehension}
We group these tasks as reading comprehension tasks. These tasks provide context information, and then the machine is asked to understand it accordingly. Thus, tasks can be format as: given the context( and other information), is the word appropriate in the context?
 \paragraph{ChID}
 ChID \citep{zheng2019chid} is a large-scale Chinese IDiom cloze test dataset, which covered from news, novels, and essays. The candidate pool contains 3,848 Chinese idioms. For each blank in the passage, there are ten candidate idioms with one golden option, several similar idioms, and others are randomly chosen from the dictionary. 
 
\paragraph{CSL}
Chinese Scientific Literature dataset contains Chinese paper abstracts and their keywords from core journals of China, covering multiple fields of natural sciences and social sciences. We generate fake keywords through tf-idf and mix them with real keywords. Given an abstract and some keywords, the task is to tell whether the keywords are all original keywords of a paper. It mainly evaluates the ability of models to judge whether keywords can summarize the document. 

\paragraph{CLUEWSC2020}
The Chinese Winograd Schema Challenge dataset
is an anaphora/coreference resolution task where the model is asked to decide whether a pronoun and a noun (phrase) in a sentence co-refer (binary classification), built following similar datasets in English (e.g., \citet{wsc} and \citet{wang2019superglue}). Sentences in the dataset are hand-picked from 36 contemporary literary works in Chinese. Their anaphora relations are then hand-annotated by linguists. 
  
For examples of each task, see Table~\ref{tab:examples}.

\begin{CJK}{UTF8}{gbsn}
\begin{table*}[!bp]
\footnotesize
\centering
\scalebox{1}{
\begin{tabular}{p{0.005\textwidth}p{0.93\textwidth}}
 \toprule
 
  \parbox[t]{1mm}{\multirow{2}{*}{\rotatebox[origin=c]{90}{{\textbf{EPRSTMT}}}}} &
\textbf{sentence:} 
外包装上有点磨损，试听后感觉不错
\\ &
\textbf{sentence (en):}
\textit{
The packaging is showing some wear, but after listening it feels good. 
}
\\ & \textbf{label:} \texttt{positive} \\
~\\
\midrule

\parbox[t]{1mm}{\multirow{2}{*}{\rotatebox[origin=c]{90}{{\textbf{CSLDCP}}}}} &
\textbf{sentence:} 
通过几年的观察和实践，初步掌握了盆栽菊花的栽培技术及方法，并进行了总结，以满足人们对花卉消费的需求，提高观赏植物的商品价值，为企业化生产的盆菊提供技术指导。
\\ &
\textbf{sentence (en):}
\textit{
After years of observation and practice, to learn and summarize the basic techniques and methods for planting the Chrysanthemum, in order to satisfy people's needs for the flowers, to increase the value of ornamental plants, and to provide technical guidance to companies on growing the Chrysanthemum. 
}
\\ & \textbf{label:} \texttt{horticulture} \\
\midrule

\parbox[t]{1mm}{\multirow{2}{*}{\rotatebox[origin=c]{90}{{\textbf{TNEWS}}}}} &
\textbf{sentence:} 
如果我的世界下架了，你会玩迷你世界吗？
\\ &
\textbf{sentence (en):}
\textit{
If Minecraft is gone, will you play miniworld?
}
\\ & \textbf{label:} \texttt{116}(news\_game) \\

\midrule
\parbox[t]{1mm}{\multirow{2}{*}{\rotatebox[origin=c]{90}{{\textbf{iFLYTEK}}}}} &
\textbf{sentence:} 
《钢铁英雄》是一款角色扮演类游戏。游戏拥有 ...... 带领他们逃出去。修复部分小错误，提升整体稳定性。
\\ &
\textbf{sentence (en):}
\textit{
"Heroes of Steel" is a role-playing game. The game has ...... all four heroes are imprisoned and you will lead them out. repair part small
Errors to improve overall stability.
}
\\ & \textbf{label:} \texttt{22}(Strategy) \\

\midrule
\parbox[t]{1mm}{\multirow{2}{*}{\rotatebox[origin=c]{90}{{\textbf{CLUEWSC}}}}} &
\textbf{text:} 
这时候放在床上枕头旁边的\underline{手机}响了，我感到奇怪，因为欠费已被停机两个月，现在\underline{它}突然响了。
\\ &
\textbf{text (en):}
\textit{
At this moment, the \underline{cellphone} on the bed next to the pillow rang. I feel this is quite strange because the cellphone plan was terminated two months ago since I did not pay the bill. Now \underline{it} was ringing all of a sudden. 
}
\\ & \textbf{label:} \texttt{true} \\

\midrule
\parbox[t]{1mm}{\multirow{2}{*}{\rotatebox[origin=c]{90}{{\textbf{CSL}}}}} &
\textbf{abst:}
不同阶段电子数据的操作都会留下表现各异的轨迹.从操作系统、计算机应用系统 ...... 分析审计电子数据轨迹在计算机系统中表现形式,可以为审计人员提供有效的审计方法
\\&
\textbf{keyword:}
[``计算机审计'', ``数据轨迹'', ``日志文件'']
\\&
\textbf{abst (en):}
\textit{
The operation of electronic data in different stages will leave different traces. From operating system, computer application system ...... provide effective audit methods for auditors by analyzing the expression of audit electronic data trace in computer system.
}\\&
\textbf{keyword (en):}
\textit{[``computer audit'', ``data trace'', ``log file'']
}
\\ & \textbf{label:} \texttt{0}(false) \\

\midrule
\parbox[t]{1mm}{\multirow{2}{*}{\rotatebox[origin=c]{90}{{\textbf{BUSTM}}}}} &
\textbf{sentence1:}
女孩子到底是不是你
\ 
\textbf{sentence2:}
你不是女孩子吗
\\&
\textbf{sentence1 (en):}
\textit{Are you a girl or not?
} \ 
\textbf{sentence2 (en):}
\textit{Aren't you a girl?
}
\\ & \textbf{label:} \texttt{paraphrase} \\

\midrule
\parbox[t]{1mm}{\multirow{2}{*}{\rotatebox[origin=c]{90}{{\textbf{OCNLI}}}}} &
\textbf{premise:}
但是不光是中国,日本,整个东亚文化都有这个特点就是被权力影响很深
\ 
\textbf{hypothesis:}
有超过两个东亚国家有这个特点
\\&
\textbf{premise (en):}
\textit{But not only China and Japan, the entire East Asian
culture has this feature, that is it is deeply influenced by
the power.
} \ 
\textbf{hypothesis (en):}
\textit{More than two East Asian countries have this feature.
}
\\ & \textbf{label:} \texttt{entailment} \\

\midrule
\parbox[t]{1mm}{\multirow{2}{*}{\rotatebox[origin=c]{90}{{\textbf{ChID}}}}} &
\textbf{content:}
中国青年报：篮协改革联赛切莫\underline{\#idiom\#}......
\\&
\textbf{candidates:} [``急功近利'', ``画蛇添足'', ``\underline{本末倒置}''(answer)]
\\&
\textbf{content (en):}
\textit{China Youth Daily: Chinese Basketball Association should not \underline{\#idiom\#} when reforming the league ......}
\\&
\textbf{candidates (en): }
\textit{[``seeking instant benefit'', ``to overdo it'', ``\underline{take the branch for the root}''(answer)]
}
\\
\bottomrule
\end{tabular}
}
\caption{Development set examples from the tasks in FewCLUE. \textbf{Bold} text represents part of the example format for each task. Chinese text is part of the model input, and the corresponding text in \textit{italics} is the English version translated from that. \underline{\textit{Underlined}} text is specially marked in the input. Text in a \texttt{monospaced font} represents the expected model output.}

\label{tab:examples}
\end{table*}
\end{CJK}

%% file: Experiments.tex
\subsection{Baseline models}\label{sec:baseline:models}

\begin{enumerate}
  \item \textbf{BERT/RoBERTa fine-tuning:} fine-tuning pre-trained models with downstream tasks is a commonly used paradigm in NLU~\citep{bert-2019,liu2019roberta}. This baseline takes pre-trained BERT/RoBERTa and fine-tunes it on each FewCLUE task.
  
  \item \textbf{Exploiting cloze questions:}
   Since most BERT-style PLMs are pre-trained with a cloze-test (also referred to as `masked language modeling', \citet{taylor1953cloze}) objective, several methods have been proposed to exploit the cloze question format, where the task of the model is to correctly predict the \texttt{[MASK]}. We experiment with two such methods.  
  
  \begin{enumerate}
    \begin{CJK*}{UTF8}{gbsn}
    \item \textbf{PET (Pattern Exploiting Training)} is a semi-supervised training scheme that reformulates the downstream task into cloze questions~\citep{schick2020small,schick-schutze-2021-PET}. Such a method includes two steps. The first step reformulates the input sentence into a cloze-style phrase containing a portion of masked tokens. For example, in text classification the input can be “这个手机壳很不错” (“This is a nice phone case”), and it’ll be reformulated into “这个手机壳很不错。我觉得[MASK]好” (“This is a nice phone case. I think it’s [MASK] good”). Here “[MASK]” is the position where the model needs to fill in to complete the classification; the second step maps the label into a single token. E.g., for sentiment classification, labels POSITIVE and NEGATIVE can mapped to the word “very”("很") and “not”("
   不") . It is not unique to map a label to a word. Many choices exist. Finally,  supervised training is performed on the  dataset by using cross-entropy loss. 

    \item \textbf{ADAPET}is an improved PET model which involves more supervised signals~\citep{tam2021-ADAPET}. It proposes two improvements over the original PET method: 
    
    (1) Decoupling Label Losses: when predicting labels in the “[MASK]” position, the model calculates the probabilities of all tokens in the vocabulary(not only the token that in candidate label) and chooses the highest-probability token as predicted label.
   
    (2) Label Conditioning: original tokens in the sentence are randomly masked, and the model tries to predict the original token based on the label. In other words, in PET the model predicts the correct label based on the input; but reversely, ADAPET predicts the correct input having the label. More specifically, given a constructed pattern with the correct label, the model is asked to predict the original token and calculate the loss; if the label is wrong, it will not calculate loss. 

    \end{CJK*}
  \end{enumerate}
  
  \item \textbf{Automatic Template Construction}
  \begin{enumerate}
      \item \textbf{LM-BFF} \citep{2021-gao-lm-bff} is a method that explores automatic prompt generation. It utilizes a generative model T5~\citep{2019t5} to automatically generate prompts for cloze questions, including template generation and label word generation, and evaluate the generated candidates based on the validation set. In addition, LM-BFF incorporates demonstrations during training to help the model better distinguish samples.
      \item \textbf{P-tuning} \citep{liu2021gpt} 
      The above methods all restrict the generated templates to be natural language. 
      P-tuning discards this assumption and proposes to learn the optimal templates by using the unused tokens in a model's vocabulary.  Only embeddings corresponding to the template tokens will be updated during training, further speeding up the training process. Models of the GPT family can also achieve good results using P-tuning.
  \end{enumerate}
  
  \item  \textbf{EFL}~\citep{wang2021-EFL} Unlike cloze-question-based models like PET and LM-BFF, Entailment as Few-shot Learners (EFL) reformulates the fine-tuning task into textual entailment and designs fine-grained textual descriptions for the labels. A basic model is obtained through training on a textual entailment dataset before fine-tuning on downstream tasks. For English, MNLI~\citep{mnli} can be used to train the model. For us, we use the machine translated version of MNLI---CMNLI~\citep{clue}. 
  
  \item \textbf{Zero-shot learning} We also adopt a standard scheme for zero-shot learning with RoBERTa and GPT3~\citep{gpt3}. Similar to PET, a prompt with masked tokens is constructed, then the model is required to predict label words at the masked position and choose the label with the highest probabilities. The difference between RoBERTa and GPT in such a scheme is that RoBERTa’s prompt can be masked in any position, while GPT can only be masked in the last position.
  
\end{enumerate}

\subsection{Implementation Details}\label{sec:implementation}

Our default baseline model, RoBERTa, is implemented using RoBERTa-wwm-ext ~\citep{cui-etal-2020-revisiting}\footnote{\urlstyle{same}\url{https://github.com/ymcui/Chinese-BERT-wwm}}. It has outstanding performance on CLUE benchmark, wide usage, and moderate size, which can be trained using around 11 GB of GPU memory. GPT model is impemented using NEZHA-Gen~\citep{wei2019nezha}\footnote{\urlstyle{same}\url{https://github.com/huawei-noah/Pretrained-Language-
Model/tree/master/NEZHA-Gen-TensorFlow}}.  
We also evaluate our benchmark on ERNIE1.0\citep{ernie1.0} to verify the benchmark.

\paragraph{PET} For PET, we use RoBERTa and ERNIE1.0 two pre-trained models, and design specific templates for each task (see our FewCLUE Github  repository\footnote{\urlstyle{same}\url{https://github.com/CLUEbenchmark/FewCLUE}}  and PaddleNLP Few-Shot-Learning\footnote{\urlstyle{same}\url{https://github.com/PaddlePaddle/PaddleNLP/tree/develop/examples/few_shot}}  Github repository for the templates).
It is noteworthy that since the input format of CHID is already cloze-style, its training using PET becomes equivalent to zero-shot learning. 

\paragraph{P-tuning} For P-tuning, our model is based on this implementation\footnote{\urlstyle{same}\url{https://github.com/bojone/P-tuning}}, with random masks on original tokens, nine prompt tokens and LSTM layers removed. We implement P-tuning method with RoBERTa, GPT and ERNIE1.0 and implement zero-shot method with both RoBERTa and GPT. 

\paragraph{EFL} For EFL, we first fine-tune the model on Chinese textual entailment task CMNLI to obtain the base model. Then, different templates are used to reformulate each task into a textual entailment task and perform downstream fine-tuning. 

\paragraph{LM-BFF}
In LM-BFF, auto-T strategy is used for prompt generation, i.e., automatic template generation combined with predefined label word mapping, as described in the original paper. A smaller beam width (30) and Chinese T5 models provided by UER\footnote{\urlstyle{same}\url{https://huggingface.co/uer/t5-base-chinese-cluecorpussmall}} \citep{zhao2019uer} are used to shorten the time for template geneartion and evaluation due to resource limit. BERT is used for fine-tuning and other hyper-parameters are the same as the original implementation. Demonstrations are not adopted since preliminary experiment shows a worse result.

\paragraph{ADAPET} For ADAPET, we also design different Chinese templates to adapt to each task.
Model is trained for 1k steps, with batch size 16, learning rate 1e-5. Other hyper-parameters follow the original paper.

For more details of our implementation such as the hyper-parameters for each method, please refer to our Github repository.\footnote{\urlstyle{same}\url{https://github.com/CLUEbenchmark/FewCLUE}}

\section{Benchmark Results}

\subsection{Human Performance}

We take human performance for those tasks available in CLUE benchmark. And for the three new tasks, human evaluation is performed in the same way as CLUE: 
We follow procedures in SuperGLUE \citep{wang2019superglue} to train the annotators before asking them to work on the test data. 
Specifically, each annotator is asked to annotate 30 to 50 pieces of data from the development set, and then compare their answers with the ground truth label. Then they are encouraged to discuss their answers with other annotators until they are confident with the task. Finally each annotator annotate 100 pieces of test data, which is used to compute the final human performance.

According to the experiment result, human achieves a final score of 82.40\%, and over 80\% in 6 out of 9 tasks. Human performance is especially high for CLUEWSC, reaching nearly 98\% of accuracy. However, humans are not satisfying in tasks containing numerous classes, e.g., about 60\% accuracy in IFLYTEK (119 classes) and CSLDCP (67 classes).

\subsection{Model Performance and Analysis}

The overall results are shown in Table~\ref{tab:main:results}. Our discoveries are summarized as follows.

\begin{table*}[t]
\scalebox{1.0}{
\small
\resizebox{\textwidth}{!}{
\begin{tabular}{l|l|lllllllll}\toprule
	\hline
    & & \multicolumn{4}{|c|}{\textbf{Single Sentence}} & \multicolumn{2}{|c|}{\textbf{Sentence Pair}} & \multicolumn{3}{|c}{\textbf{MRC}} \\
    \hline
    \midrule
  Method & Score & EPRSTMT & CSLDCP & TNEWS & IFLYTEK & OCNLI & BUSTM & CSL & CHID & WSC \\ \midrule
\hline
Majority & 29.04& 50.0 & 1.5 & 6.7 & 0.8 & 38.1 & 50.0 & 50.0 & 14.3 & 50.0 \\\hdashline
Human & 82.50 & 90.0 & 68.0 & 71.0 & 66.0 & 90.3 & 88.0 & 84.0 & 87.1 & 98.0 \\\hdashline
FineTuningR & 44.10 & 65.4(7.7)  & 35.5(2.5) & 49.0(1.6) & 32.8(1.7) & 33.0(0.34) & 60.7(9.1) & 50.0(0.1) & 14.9(0.4) & 55.6(14)  \\\hdashline
 Zero-shotR & 44.60 & 85.2 & 12.6 & 25.3 & 27.7 & 40.3 & 50.6 & 52.2 & 57.6 & 50.0 \\
Zero-shotG & 43.40 & 57.5 & 26.2 & 37.0 & 19.0 & 34.4 & 50.0 & 50.1 & \textbf{65.6} & 50.3 \\\hdashline

PET & 57.44 & 86.7(1.0) & 51.7(1.0) & 54.5(1.2) & 46.0(1.1) & 44.0(0.4) & 56.0(5.0)  & 59.4(1.3) & 61.2(1.1) & 57.5(2.7) \\\hdashline

LM-BFF & 56.32 & 85.6 (0.9) & 54.4 (3.1) & 53.0 (2.1) & 47.1 (2.6) & 41.6 (4.0) & 57.6 (3.4) & 51.7 (2.4) & 61.2 (1.10) & 54.7 (6.7) \\
P-tuningR &  \textbf{59.91} & \textbf{88.3}(0.7) & 56.0(1.1) & 54.2(1.0) & \textbf{57.6}(0.9) & 41.9(1.9) & 60.9(2.9) & \textbf{62.9}(2.3) & 59.3(1.4) & 58.1(2.2)  \\\hdashline
EFL & 55.91 & 84.9(0.4) & 45.0(2.3) & 52.1(0.8) & 42.7(1.1) & \textbf{66.2}(1.4) & \textbf{71.8}(0.8) & 56.6(1.8) & 30.9(1.9) & 53.0(3.1) \\\hdashline

FineTuning$_{ernie1.0}$ & 48.34 & 66.5(6.3) & 57.0(3.4) & 51.6(3.3) & 42.1(3.7) & 32.0(1.6) &   60.4(5.7) & 60.1(3.5) & 15.0(0.4) & 50.3(1.4)\\

PET$_{ernie1.0}$ & 56.39 & 84.0(4.8) & \textbf{59.9}(3.0) & \textbf{56.4}(0.8) & 50.3(2.8) & 38.1(0.3) &   58.4(3.4) & 61.1(4.4) & 40.6(2.5) & \textbf{58.7}(1.4)\\

P-tuning$_{ernie1.0}$ & 54.37 & 80.6(5.3) & 56.6(6.6) & \ 55.9(1.3) & 52.6(5.0)  &  35.7(0.8) & 60.8(5.3) & 51.8(3.5) & 39.6(1.7)  & 55.7(2.0) \\

EFL$_{ernie1.0}$ & 52.27 & 76.7(6.3) & 47.9(2.3) & 56.3(0.6) & 52.1(2.2) & 48.7(2.5) & 54.6(8.1)  & 52.8(3.2) & 30.3(4.3) & 52.3(3.7) \\
\hline
\bottomrule

\end{tabular}}
}

\caption{Main results of different learning mechanisms on FewCLUE. Single sentence task: EPRSTMT, CSLDCP, TNEWS and IFLYTEK. Sentence-pair task: BUSTM, OCNLI. MRC task: CHID, WSC, CSL. Majority: majority class. B, R and G stands for BERT, RoBERTa and GPT series models respectively.  MRC: Machine reading comprehension requires a machine to answer questions based on a given context. As these tasks all have context, so that  can be regarded as MRC tasks. \label{tab:main:results}
} 
\end{table*}

\paragraph{Model potentials:} The best model with the highest overall score (59.91) is far lower than human performance (82.50), nearly 23 points lower, which shows that few-shot learning still have huge potentials.
\paragraph{New learning paradigm:} recently proposed few-shot learning methods show a remarkable improvement compared to traditional fine-tuning. For example, with the same pre-trained model, PET is about 18 points higher than fine-tuning. Especially in EPSTMT, PET is 24\% higher, while zero-shot learning is also 21\% higher than fine-tuning. The reason might be the insufficient training on significant trainable parameters during fine-tuning using very limited samples. Besides, PET, P-tuning and the zero-shot method are also substantially better than fine-tuning on CHID. It might because CHID is natively suitable for cloze-style learning methods.
    
    
\paragraph{Comparing few-shot learning methods:} 
We have implemented five few-shot learning methods, including PET, ADAPET, LM-BFF, Ptuning, EFL. The first four methods transformed the task to Cloze task, the last one transforming the task to natural language inference. Among the first four methods, ADAPET and Ptuning, try to find template automatically, but in a differnt way. PET achieves better results than other methods. P-tuningG, stands for P-tuning using a generated-based model, it performs the worst in our experiment. EFL achieves quite good results for natural language inference tasks, OCNLI and BUSTM, but performs poorly on CHID, a machine learning comprehension task.

\paragraph{Comparison of different pre-trained models on few-shot learning methods:} 
We compared the effects of RoBERTa and ERNIE1.0 on the 3  few-shot learning methods of PET, P-tuning and EFL. As shown in Table 2: The pre-training model has a great influence on the different FewCLUE tasks. For example, For the PET method, the effect of ERNIE1.0 on the tasks of EPRSTMT, OCNLI and CHID is obviously worse than that of RoBERTa, and the effect on the tasks of CSLDCP, TNEWS and IFLYTEK is obviously better than that of RoBERTa. for the EFL method, the effect of ERNIE1.0 on EPRSTMT, OCNLI, BUSTM and CSL tasks is significantly worse than that of RoBERTa, and the effect on CSLDCP, TNEWS and IFLYTEK tasks is obviously better than RoBERTa.  Overall, the average effect of RoBERTa on the three methods of PET, P-Tuning, and EFL is better than that of ERNIE1.0. In terms of specific tasks, the two pre-training models have their own results.

\paragraph{Zero-shot learning capability:} zero-shot learning methods have obtained good results in some tasks without any training data. For example, in IFLYTEK, zero-shot models achieve an accuracy of 27.7\%,  which is only 5\% lower than the fine-tuning model, while random guessing has only 1\% accuracy. Similar discoveries also show up in CSLDCP. The improvement on single-sentence tasks is more significant than sentence-pair tasks. It might be because most of the models' knowledge still comes from pre-training phase, and the input of single-sentence tasks is more similar to what they have seen in the pre-training corpus. However, for sentence-pair and reading comprehension tasks, the input’s consistency is likely to be damaged whether using manual or auto-generated templates, which leads to less improvement.
   
\paragraph{Learning instability for single train/dev set:} we discover that there is a severe fluctuation among the results produced by different training sets. Also, models that achieve the best validation result do not necessarily lead to the best test result. We assume that this is because the training and validation splits are too small to represent the overall distribution of the whole dataset. To show this phenomenon, we perform three comparison experiments based on PET and TNEWS datasets. For each experiment, training and validation sets are randomly sampled again from the full dataset, while test set is the same as above.

\begin{enumerate}
    \item 20 samples/class, training/validation ratio 1:1.
    \item 20 samples/class, training/validation ratio 1:3.
    \item 7 samples/class, training/validation ratio 1:1.
\end{enumerate}

Three experiments obtain the accuracy of 50.8\%, 51.2\%, 50.0\% using the best validation model, which are sub-optimal compared with 53.7\%. We also discover that validation results for exp. 2 have higher variance than exp. 1 and 3. It shows that accuracies are quite unstable using a single validation set, and it is also hard to obtain the best testing accuracy. Therefore, how to improve the stability of few-shot learning is a valuable research topic in the future.


\subsection{Task Analysis}

We have found that in few-shot learning, the difficulties of tasks for humans and models are substantially different. On the one hand, tasks like coreference resolution WSC are extremely easy for humans (98\%), but it is hard for models to perform, which is no different from random guessing ($\sim$50\%); However, in some other tasks, the gap is much smaller. For CSLDCP, a task with 67 labels, humans only archive an accuracy of 68\%, which is relatively low compared to other tasks, while PET’s results are only 16 points lower.  Thus, we can anticipate that few-shot models still have space to progress.

%% file: Conclusion.tex
In this paper, we introduced FewCLUE, the first systematic and comprehensive few-shot learning benchmark in Chinese. 
Our benchmark consists of nine tasks, including single-sentence, multi-sentence, and machine reading comprehension tasks. 
We make the first attempt to conduct a fair and comprehensive experiment across five state-of-the-art few-shot learning models on our newly proposed benchmarks.
In addition, we release all code and benchmarks to facilitate future benchmarking, research, and model development.